\documentclass{article}
\usepackage{spconf,amsmath,graphicx}
\usepackage{calrsfs}
\usepackage{calc}
\usepackage{amsmath,amssymb,pifont}
\usepackage{multirow}

\DeclareMathAlphabet{\pazocal}{OMS}{zplm}{m}{n}
\newcommand{\D}{\pazocal{D}}

\renewcommand{\thefootnote}{\alph{footnote}}

\newcommand{\astfootnote}[1]{
	\let\oldthefootnote=\thefootnote
	\setcounter{footnote}{0}
	\renewcommand{\thefootnote}{\fnsymbol{footnote}}
	\footnote{#1}
	\let\thefootnote=\oldthefootnote
}

\title{Listening while Speaking and Visualizing: \\Improving ASR through Multimodal Chain}
\name{Johanes Effendi$^{1,2}$, Andros Tjandra$^{1}$, Sakriani Sakti$^{1,2}$, Satoshi Nakamura$^{1,2}$}
\address{
	$^1$Nara Institute of Science and Technology, Japan\\
	$^2$RIKEN, Center for Advanced Intelligence Project AIP, Japan\\
}

\begin{document}
	%
	\maketitle
	\begin{abstract}
Previously, a machine speech chain, which is based on sequence-to-sequence deep learning, was proposed to mimic speech perception and production behavior. Such chains separately processed listening and speaking by automatic speech recognition (ASR) and text-to-speech synthesis (TTS) and simultaneously enabled them to teach each other in semi-supervised learning when they received unpaired data. Unfortunately, this speech chain study is limited to speech and textual modalities. In fact, natural communication is actually multimodal and involves both auditory and visual sensory systems. Although the said speech chain reduces the requirement of having a full amount of paired data, in this case we still need a large amount of unpaired data. In this research, we take a further step and construct a multimodal chain and design a closely knit chain architecture that combines ASR, TTS, image captioning, and image production models into a single framework. The framework allows the training of each component without requiring a large number of parallel multimodal data. Our experimental results also show that an ASR can be further trained without speech and text data and cross-modal data augmentation remains possible through our proposed chain, which improves the ASR performance.

\end{abstract}

\vspace{-0.7cm}
\section{Introduction}
\vspace{-0.3cm}

Researchers have been working in speech recognition technology for many decades. State-of-the-art ASR systems are currently based on end-to-end deep learning frameworks. Traditionally, they are usually trained by applying supervised learning techniques that rely on the availability of speech data and corresponding transcriptions. To improve the performance in the presence of unexpected acoustic variability, we usually collect more data to train more detailed models. Although some systems have successfully reached parity with humans \cite{humanparity_mic,humanparity_ibm}, such a learning style can only be perfectly fit for recognizing the speech of about 10-20 of the world's most common languages. For many others, the problem is the size of the required speech and corresponding transcriptions are usually unavailable.

Recently, approaches that utilize learning from source-to-target and vice-versa as well as feedback links have gained attention and provide the possibility of training models with unpaired datasets. He et al. \cite{he2016dual} and Cheng et al. \cite{cheng_robust}, recently published work that addressed a mechanism called dual learning in neural machine translation (NMT). Their system has a dual task: source-to-target language translation (primal) versus target-to-source language translation (dual). It can leverage monolingual data to improve neural machine translation. In image processing, several methods have also been proposed to achieve unsupervised joint distribution matching without any paired data, including DiscoGAN \cite{discogan}, CycleGAN \cite{CycleGAN2017}, and DualGAN \cite{dualgan}. The framework provides learning to translate an image from a source domain to a target domain without paired examples based on a cycle-consistent adversarial network. Implementation on voice conversion applications has also been investigated \cite{vc_cyclegan}. However, most only work with the same domain between the source and the target.

In speech-language processing, the speech chain framework \cite{tjandra_schain1,tjandra_schain2,tjandra_chain3} was proposed to integrate human speech perception and production behaviors that utilize the primal model (ASR) that transcribes a text, given the speech versus the dual model (TTS) that synthesizes the speech given the text. Perhaps this is the first framework that was constructed on a different domain (speech versus text). The approach provides freedom from needing a large amount of speech-text paired data and possibilities to improve ASR performance in semi-supervised learning by allowing ASR and TTS to teach each other, given only text or only speech data. Unfortunately, although this approach reduces the requirement of a large amount of paired data, we still need a large amount of unpaired data. Furthermore, this study is limited to speech and textual modalities. In fact, natural communication is actually multimodal that involves both auditory and visual sensory systems.

\begin{figure*}[!h]
	
	\vspace{-0.7cm}
	\centering
	\includegraphics[width=1.7\columnwidth]{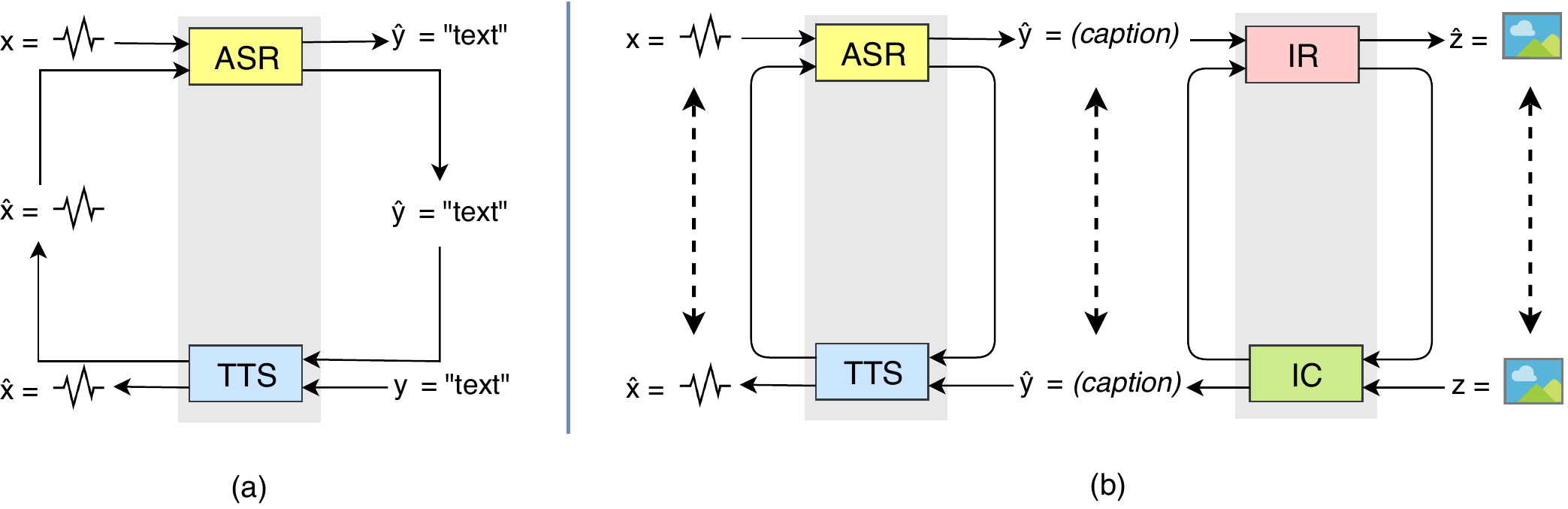}
	\vspace{-0.4cm}
	\caption{Architecture: (a) speech chain framework \cite{tjandra_schain1}, and (b) our proposed multimodal chain mechanism.}
	\label{fig:multimodal_chain_fig}
	\vspace{-0.5cm}
\end{figure*}

In this research, we constructed the first framework that accommodates triangle modality (speech, text, and image) and addressed the problems of speech-to-text, text-to-speech, text-to-image, and image-to-text. Our new framework mimics the mechanism of the entire human communication system with auditory and visual sensors. Similar to a machine speech chain, it allows each component to be trained without needing a large number of parallel multimodal data. In addition to the above advantages, through closely knit chain architecture that combines ASR, TTS, IC, and IR models into a single framework, it further frees us from needing a large amount of unpaired data. For example, specifically to ASR tasks, even when no more speech and text data are available, it is still possible to improve ASR by cross-modal data augmentation through our proposed chain.

\vspace{-0.2cm}
\section{Multimodal Chain Framework}
\label{sec:mmchain}
\vspace{-0.3cm}
Figure~\ref{fig:multimodal_chain_fig} illustrates (a) the original speech chain framework \cite{tjandra_schain1} and (b) our proposed multimodal chain mechanism. In this extension, we included image captioning (IC) and image retrieval (IR) models to incorporate visual modality into the chain. The framework consists of dual loop mechanisms between the speech and visual chains that involve quadruple learning components: ASR, TTS, IC, and IR. In the speech chain, sequence-to-sequence ASR and sequence-to-sequence TTS models are jointly trained in a loop connection, and in the visual chain, neural image captioning and neural embedding-based image retrieval models are also jointly trained in a loop connection. Both chains (speech and visual components) are allowed to collaborate by text modality. 

The sequence-to-sequence model in closed-loop architecture allows us to train our entire model in a semi-supervised fashion by concatenating both the labeled and unlabeled data. To further clarify the learning process, we describe the mechanism based on the availability condition of the training data:

\begin{enumerate}
	\item \textbf{Paired speech($x$)-text($y$)-image($z$) data exist: separately train ASR, TTS, IC and IR (supervised learning)}\\
	Given complete multimodal dataset $\D^{P_{xyz}}$, we can set-up speech utterances $x$ and corresponding text transcriptions $y$ as dataset $\D^{P_{xyz}}_{xy}$ to separately train both the ASR and TTS models in a supervised manner. ASR losses $L^{P}_{ASR}$ and $L^{P}_{TTS}$ are calculated directly with teacher-forcing, where the ground truth for each time step is used as input when decoding. We can also set-up images $z$ and captions $y$ as dataset $\D^{P_{xyz}}_{yz}$ to separately train the IC and IR models with supervised learning. The IC model is trained with teacher-forcing on reference caption $y$, and the IR model is trained with pairwise rank loss \cite{westonrankloss} on reference image $z$ and its contrastive sample.	
	\item \textbf{Unpaired speech ($x$), text ($y$), images ($z$) data exist: jointly Train ASR\&TTS in the speech chain and IC\&IR in the visual chain (unsupervised learning)}\\
	In this case, although speech, text, and image data are available, they are unpaired. 
	\begin{enumerate}
		\item \textbf{Only using speech data: unrolled process ASR$\rightarrow$TTS in a speech chain}\\
		Here we only use speech utterances $x$ of dataset $\D^{U_{xyz}}$, and ASR generates transcription $\hat{y}$ for TTS to reconstruct. Reconstructed transcriptions $\hat{x}$ calculate loss $L^{U}_{TTS}$ between $x$ and $\hat{x}$ and update the model parameter.
		\item \textbf{Only using image data: unrolled process IC$\rightarrow$IR in a visual chain}\\
		Using only image $z$ in dataset $\D^{U_{xyz}}$, image captions $\hat{y}$ are generated with the IC model. These captions are then used by the IR model to update its multimodal space using pairwise rank loss, which resulted in loss $L^{U}_{IR}$.
		\item \textbf{Only using text data: unrolled process TTS$\rightarrow$ASR in the speech chain and IR$\rightarrow$IC in the visual chain}\\
		Given only the text in dataset $\D^{U_{xyz}}$, TTS generates speech utterance $\hat{x}$ for the ASR, which then reconstructs the speech utterances into text $\hat{y}$ in which reconstruction loss $L^{U}_{ASR}$ between $y$ and $\hat{y}$ can be calculated. On the other hand, image captions $y$ retrieve images $\hat{z}$, which are reconstructed into text $\hat{y}$ using the IC model in which losses $L^{U}_{IC}$ are calculated between $y$ and $\hat{y}$.
	\end{enumerate}

	\item \textbf{Single data (either speech ($x$), text ($y$), or images ($z$)) exist: train ASR \& TTS jointly in the speech chain and IC \& IR in the visual chain (unsupervised learning)}\\
	In this case, only a single modality data (either speech, text, or image) is available, and the others are empty. 
	
	\begin{enumerate}
		\item \textbf{Only text data exist: train the speech and visual chains, as in 2(c)}\\
		If only text data are available in dataset $\D^{U_y}$, we can separately perform unrolled process TTS$\rightarrow$ASR in the speech chain and IR$\rightarrow$IC in the visual chain
		\item \textbf{Only speech data exist: speech chain $\rightarrow$ visual chain}\\
		If only speech data are available in dataset $\D^{U_x}$, first we perform unrolled process ASR$\rightarrow$TTS in the speech chain (See 2(a)). The generated text transcription $\hat{y}$ is then used to performs IR$\rightarrow$IC in the visual chain (See 2(c)). 
		\item \textbf{Only image data exist: visual chain $\rightarrow$ speech chain}\\
		If only image data are available in dataset $\D^{U_z}$, first we perform unrolled process IC$\rightarrow$IR in the visual chain (See 2(b)). The generated image caption $\hat{y}$ is then used to perform unrolled process TTS$\rightarrow$ASR in the speech chain (See 2(c)).
	\end{enumerate}
\end{enumerate}

Here our main concern is the last point (3(c)). We are interested to learn whether in the situation where only image data exist (no more speech and text data are available) we can still improve the ASR performance through a learning process from a visual chain to a speech chain by leveraging cross-modal data augmentation.

We combine all of the losses and update both the ASR and TTS models as well as the IC and IR models:

\scriptsize{
	\begin{equation}
	\begin{split}
	L_{sc} = &\alpha_{ASR} L^{P}_{ASR} + \alpha_{TTS} L^{P}_{TTS} + \\
	&\beta_{ASR} L^{U}_{ASR} + \beta_{TTS}L^{U}_{TTS} \\
	\end{split}
	\end{equation}
	\vspace{-0.4cm}
	\begin{align}
	\theta_{ASR} = &Optim(\theta_{ASR}, \bigtriangledown_{\theta_{ASR}}L) \\
	\theta_{TTS} = &Optim(\theta_{TTS}, \bigtriangledown_{\theta_{TTS}}L)
	\end{align}
	
	\vspace{-0.5cm}
	
	\begin{gather}
	L_{vc} = \gamma_{IC} L^{P}_{IC} + \gamma_{IR} L^{P}_{IR} + \delta_{IC} L^{U}_{IC} + \delta_{IR} L^{U}_{IR} \\
	\theta_{IC} = Optim(\theta_{IC}, \bigtriangledown_{\theta_{IC}}L) \\
	\theta_{IR} = Optim(\theta_{IR}, \bigtriangledown_{\theta_{IR}}L) 
	\end{gather}
	
}%

\normalsize
\noindent
which results in losses $L_{sc}$ and $L_{vc}$ for the speech and visual chains. Parameters $\alpha$, $\beta$, $\gamma$, and $\delta$ are hyper-parameters for scaling the loss between the supervised (paired) and unsupervised (unpaired) losses in each chain.

\vspace{-0.2cm}
\section{Multimodal Chain Components}

\vspace{-0.3cm}
In this section, we briefly describe all of the components inside the multimodal chain framework.

\vspace{-0.3cm}
\subsection{Sequence-to-sequence ASR}
\vspace{-0.2cm}
\label{ssec:asr}
We use the sequence-to-sequence ASR model with attention, whose architecture resembles a previous one used in \cite{tjandra_schain1} that is also based on LAS framework \cite{chan2016listen}. It directly models conditional probability $P(y|x)$ of transcription $y$ given speech feature $x$. For the speech feature, the MFCC or the mel-spectogram are usually encoded by a bidirectional LSTM encoder. The hidden representations are then attended by a LSTM or GRU decoder that decodes a sequence of characters or phonemes.

\subsection{Sequence-to-sequence TTS}
\label{ssec:tts}
A sequence-to-sequence TTS is a parametric TTS that generates sequence of speech feature $x$ from transcription $y$. We also used similar architecture as a previous one used in \cite{tjandra_schain1} which is based on a slight modification to Tacotron \cite{wang_taco}. Tacotron produces a mel-spectogram given the text utterances, and is further transformed into a linear spectogram so that the speech signal can be reconstructed using the Griffin-Lim algorithm \cite{griffin_algo}.

\subsection{Image Captioning}
\label{ssec:ic}
\begin{figure}[!h]
	\vspace{-0.3cm}
	\centering
	\includegraphics[width=0.9\columnwidth]{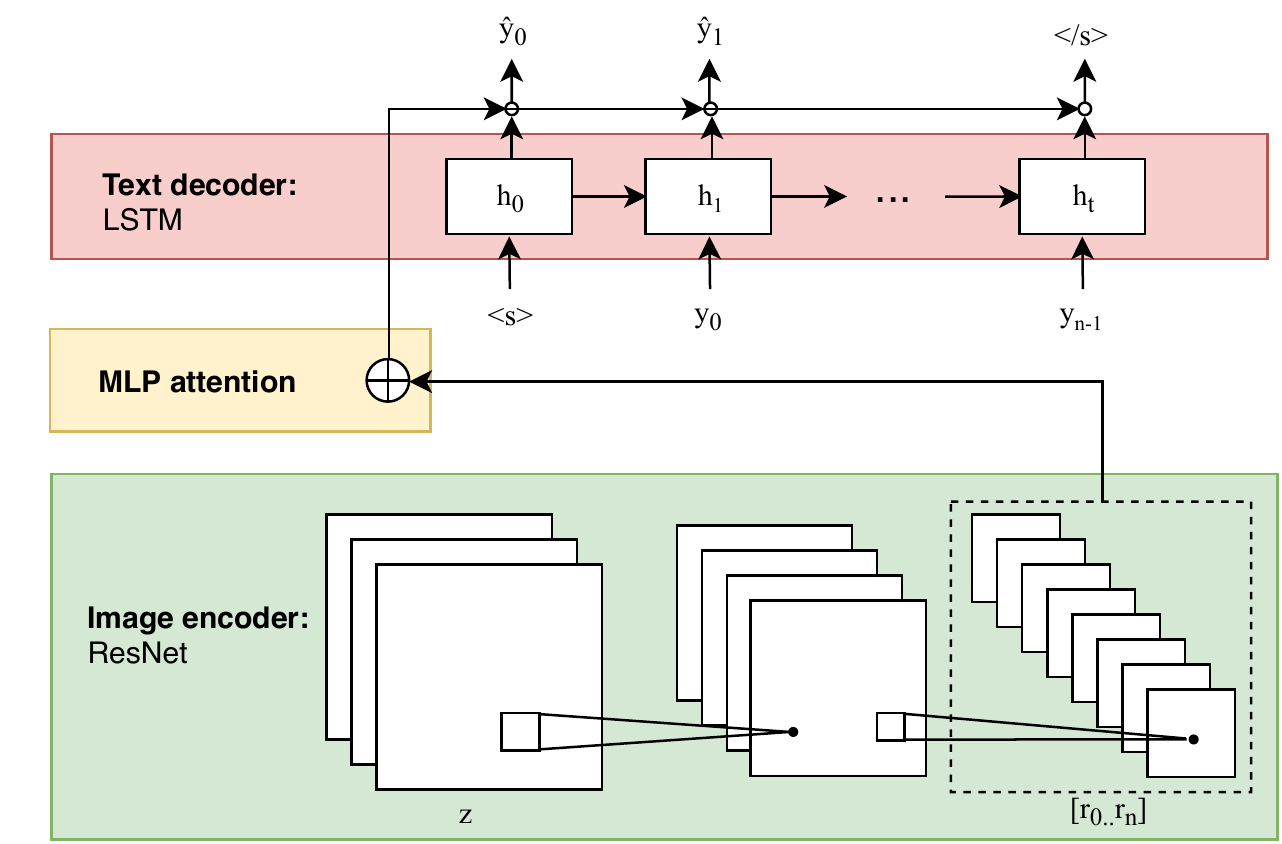}
	\caption{Caption model}
	\label{fig:caption_fig}
\end{figure}
An image captioning model receives image $z$ and produces caption prediction $\hat{y} = [\hat{y}_0..\hat{y}_n]$ by minimizing softmax cross entropy loss against original caption $y = [y_0..y_n]$. We utilized similar architecture as \cite{xu2015show}, where image $z$ are encoded through a series of convolutional neural network $enc_{img}$ resulting in a high level feature representation within a certain number of region $enc_{img}(z) = [r_0..r_n]$ that represent parts of the image. During decoding the $[\hat{y}_0..\hat{y}_n]$, linear attention grounds each decoded word into correlated image region $r_n$ by calculating alignment probability $a_t(enc_{img}(z)) = Align([r_0..r_n], h_t)$ over decoder states ${h_t}$. Unlike Xu et al's model, we use ResNet \cite{he2016deep} instead of VGG \cite{simonyan2014very}.
\vspace{-0.3cm}
\subsection{Image Retrieval}
\label{ssec:ir}
\begin{figure}[!h]
	\centering
	\includegraphics[width=0.9\columnwidth]{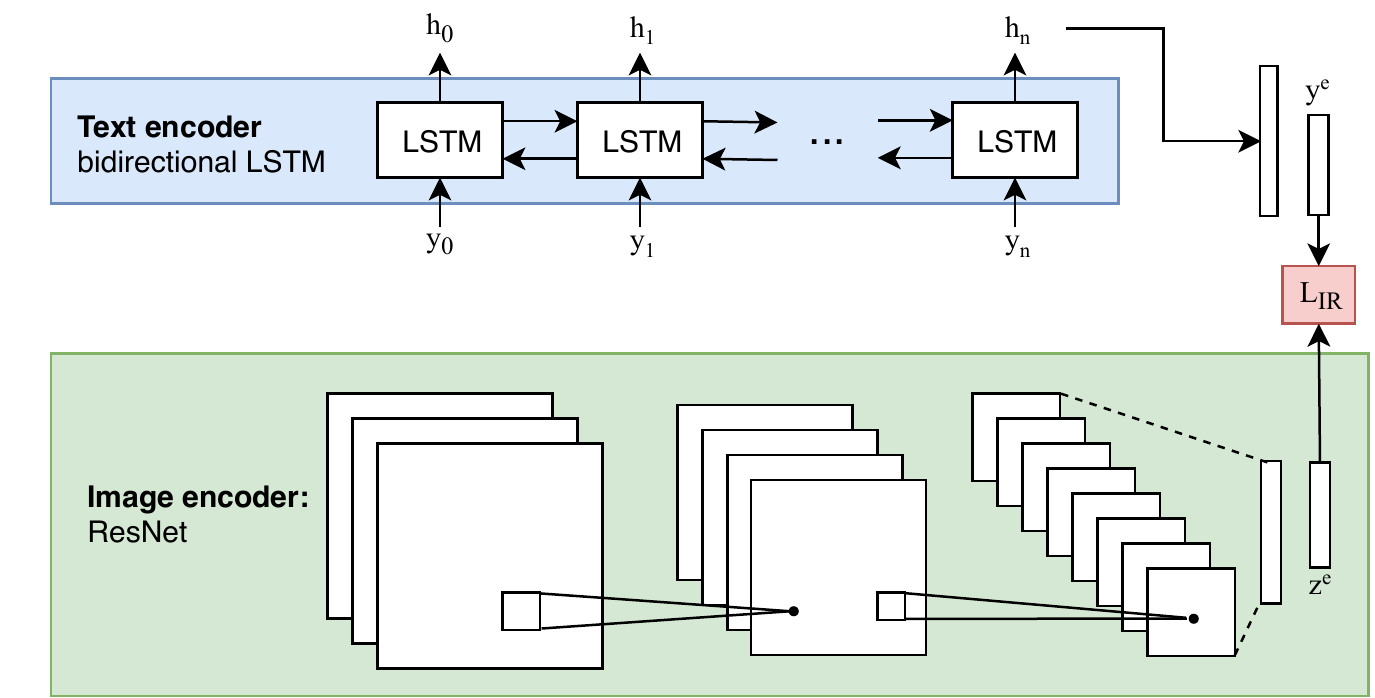}
	\caption{Retrieval model}
	\label{fig:retrieve_fig}
	\vspace{-0.3cm}
\end{figure}

Neural IR models \cite{dongw2vv,vilaltastudying,ma_mmconv,calixto_mlmme} are implemented by realizing a multimodal embedding between image $z$ and its caption $y$. Image embedding $z^e$ is usually extracted from a series of pretrained convolutional neural networks followed by linear transformation. Recurrent neural network encoder are used to generate caption embedding $y^e$, which is transformed from the last encoder state $h_n$ that encodes sequence of words $[y_0..y_n]$. These two embedding representations are then trained using pairwise rank loss function $L_{IR}$ to join them into a unique multimodal embedding space.

As shown in Eq. \ref{eq:prl}, this procedure reduces mean squared distance $d$ between each image embedding $z^e$ with related caption represented with text embedding $y^e$, and increases its distance with unrelated caption $\hat{y}^e$. Margin $M$ is used to distance the already similar pairs, providing space to optimize the hard-positive examples:
\begin{equation}
\label{eq:prl}
\small
\begin{split}
L_{IR} = \sum_{|y^e|} \sum_{|\hat{z}^e|} max\{0, M - d(y^e,z^e) + d(y^e,\hat{z}^e) \} + \\
\sum_{|z^e|} \sum_{|\hat{y}^e|} max\{0, M - d(z^e,y^e) + d(z^e,\hat{y}^e) \}
\end{split}
\end{equation}

\section{Experimental Set-Up}
\subsection{Corpus Dataset}
\vspace{-0.5cm}
\begin{table}[h]
	\vspace{-0.3cm}
	\small
	\centering
	\label{tab:data-composition}
	\caption{Training data partition for Flickr30k with three conditions: (1) available paired data are denoted as $\bigcirc$, (2) available data but unpaired are denoted as $\blacktriangle$, and unavailable data are denoted as $\times$.}
	\vspace{0.3cm}
	\begin{tabular}{|c|c|c|c|c|c|}
		\hline
		\multirow{2}{*}{\textbf{Dataset}} & \textbf{Speech}   & \textbf{Text}    & \textbf{Image}   & \multirow{2}{*}{\textbf{\# Data}\textsuperscript{*}}& \textbf{Training} \\
		 & \textbf{$x$}      & \textbf{$y$}     & \textbf{$z$}     & & \textbf{Type} \\ \hline \hline
		$D^{P_{xyz}}$    & $\bigcirc$        & $\bigcirc$       & $\bigcirc$       & 2000             & 1 (paired)   \\ \hline
		$D^{U_{xyz}}$    & $\blacktriangle$  & $\blacktriangle$ & $\blacktriangle$ & 7000             & 2 (unpaired) \\ \hline
		$D^{U_x}$        & $\blacktriangle$  & $\times$         & $\times$         & 10000            & 3 (unpaired) \\ \hline
		$D^{U_z}$        & $\times$          & $\times$         & $\blacktriangle$ & 10000            & 3 (unpaired) \\ \hline
	\end{tabular}
	\vspace{-0.2cm}
\end{table}

\interfootnotelinepenalty=0
\let\thefootnote\relax\footnotetext{\textsuperscript{*}Different modality has different unit (speech=utterance, text=sentence, image=picture). In paired data, one image is at least associated with one text sentence and one speech utterance.}

We ran our experiment with the Flickr30k dataset \cite{plummer_2015} that has 31,014 photos of everyday activities, events, and scenes. Similar to other image captioning datasets, each image has five captions with a vocabulary of 18k words. However, since we use this dataset not only for captioning but also for retrieval, we need to maintain a balance between source and target. For image captioning, we use one caption per image to make the learning target consistent by avoiding one-to-many confusion. Conversely, in image retrieval we used all five captions because the learning target is already consistent. To train the speech counterpart of our proposed architecture, we generated speech from the Flickr30k captions using single speaker Google TTS. The transcription resulted in 145k utterances, all with the same speaker, with the total duration of 178.8 hours. Each caption typically consists of 12.32 words in the form of a sentence.

In our experiment, we used the dev and test set of Flickr30k for validation and testing, respectively. Similar to the training data, Google TTS is also used to generate the corresponding speech (single speaker voice) of these two sets.

To show the capability of our model for semi-supervised learning, we formulated the training part of the dataset into four parts. For more details, see the specifications in Section \ref{sec:mmchain}. The selection of which data belongs to which part was done randomly since all the training data are shuffled before being split into four parts. The first part, $D^{P_{xyz}}$, was used to supervisedly train each model (\textbf{Type 1}), because all the data are paired ($\bigcirc$). Next, the $D^{U_{xyz}}$ dataset, which has all the modalities available but unpaired ($\blacktriangle$) was used to separately train the speech  and visual chains (\textbf{Type 2}). $D^{U_x}$ and $D^{U_z}$ are assumed to be a single modality corpus ($\times$), which only has speech and images without any transcription or captioning. By decoding the $D^{U_x}$ dataset into image captions, and $D^{U_z}$ into utterance transcriptions, we can use the generated data to further semi-supervisedly improve each model (\textbf{Type 3}). 

Without our proposed architecture, these monomodal data $D^{U_x}$ and $D^{U_z}$ cannot be used because their modality is completely unrelated to the chain in the other modality pair. As mentioned above, our main concern is to know whether it is still possible to improve ASR performance in the \textbf{Type 3} situation, where only image data is available on $D^{U_z}$. 




\vspace{-0.2cm}
\subsection{Model Details}
\vspace{-0.2cm}
We used a standard sequence-to-sequence model with an MLP attention module for ASR as mentioned in Section \ref{ssec:asr}. For the TTS, we used Tacotron \ref{ssec:tts}. The features and architecture of the ASR and TTS models are similar with Tjandra et al. (2017) single speaker speech chain models \cite{tjandra_schain1}. We used an Adam optimizer with a learning rate of 1e-3 for the ASR model, 2.5e-4 for the TTS model, and 1e-4 for the IC model. For the IR model, we used a stochastic gradient descent with a 0.1 learning rate. 

In the visual chain, we implemented the IC and IR models as previously described in Sections \ref{ssec:ic} and \ref{ssec:ir}. For the convolutional part that extracts the image features, we used ResNet \cite{he2016deep} as an image encoder in IC and IR. In the IC model, we removed the last two layers of the ResNet, which yields a 14x14 latent representation of the image region in which the decoder could attend to. Then, for the IR model, we removed the last layer, giving us a 2048-dimensional hidden representation that can be regarded as an image representation. These representations are then linearly transformed into 300-dimensional image embeddings. On the other hand, we generated text embeddings using a single-layer bidirectional LSTM with 256 hidden sizes in each direction.

We decoded the transcription in the speech chain using beam-search decoding with a size of three. Similarly, during the visual chain operation, the IC model produced its hypothesis using beam-search decoding of the same size. The granularity difference between each chain (i.e., char and word) was fixed during training between the chains. To simulate sampling in the IR hypothesis, we randomly sampled one hypothesis from five candidates.

\section{Experiment Results}
\subsection{A Large Amount of Paired Data Exists - Topline Case}
\vspace{-0.4cm}
\begin{table}[h]
	\vspace{-0.3cm}
	\centering
	\caption{Our ASR and TTS performance in comparison with the existing published results\label{tab:topasrtts}}
	\begin{tabular}{|l|c|c|}
		\hline
		\multicolumn{1}{|c|}{\multirow{2}{*}{\textbf{Data}}} & \textbf{ASR} & \textbf{TTS} \\
		 & \textbf{CER(\%)} & \textbf{L2-norm$^{2}$} \\ \hline
		Kim et al. \cite{kim2017joint}  & 11.08 & - \\ \hline
		Tjandra et al. \cite{tjandra2018multi} & 6.60 & 0.682 \\ \hline
		Ours  & 6.87 & 0.653 \\ \hline
	\end{tabular}
\end{table}
\vspace{-0.7cm}
\begin{table}[h]
	\centering
	\caption{Our IC and IR performance in comparison with the existing published results\label{tab:topicir}}
	\begin{tabular}{|l|c|c|c|}
		\hline
		\multicolumn{1}{|c|}{\multirow{2}{*}{\textbf{Data}}} & \textbf{IC} & \multicolumn{2}{|c|}{\textbf{IR}} \\
		 & \textbf{BLEU1} & \multicolumn{1}{|c|}{\textbf{R@10$\uparrow$}} & \textbf{med r$\downarrow$} \\ \hline
		Xu et al. (2015) \cite{xu2015show} & 67.00 & - & -                            \\ \hline
		Vilalta et al. (2017) \cite{vilaltastudying} & - & 59.8 & 6                            \\ \hline
		Ours & 66.27 & 62.42 & 5                            \\ \hline
	\end{tabular}
	\vspace{-0.3cm}
\end{table}
\interfootnotelinepenalty=10000
\let\thefootnote\relax\footnotetext{\textsuperscript{*}We trained our baseline model with the only 2k paired data to simulate a real-condition where an only small amount of paired dataset is available to show that our chain can improve the initial model that was only trained with a small amount of data using semi-supervised learning fashion.}

In this subsection, we assumed that a large amount of paired data exists. Therefore, we can train each model independently using supervised training. Here we compared the performance of all the system components with the existing published results on a well-known dataset. For the ASR and TTS tasks, we evaluated the performance of our models on the Wall Street Journal dataset \cite{wsj_corpus}, which is a natural multispeaker speech corpus. Our settings for the training, development, and test sets are identical as the Kaldi s5 recipe \cite{povey11asru}. We trained our model with the WSJ-SI284 data. Our validation set was dev93, and our test set was eval92. For the IC and IR tasks, we evaluated the performance of our models on a full set of Flickr30k.

Tables \ref{tab:topasrtts} and \ref{tab:topicir} show a comparable evaluation for the ASR-TTS and IC-IR tasks. As seen in the table, our ASR performs better than Kim et al. \cite{kim2017joint} and provides a similar performance to Tjandra et al. \cite{tjandra_schain2}. Our TTS model also performs on par with Tjandra et al. \cite{tjandra_schain2}. Our IC model also performed on par on the Flickr30k dataset with the work by Xu et al. by a 0.7 BLEU1 margin. Finally, we compared our IR model with Vilalta et al. (2017) who proposed a full network embedding model in which our model performed slightly better. These results reveal that in a fully supervised scenario, our model works as well as previously published papers.

\begin{table}[h]
	\vspace{-0.3cm}
	\centering
	\caption{ASR and TTS performance using Multimodal Chain\label{tab:asrtts}}
	\begin{tabular}{|l|c|c|}
		\hline
		\multicolumn{1}{|c|}{\multirow{2}{*}{\textbf{Data}}} & \textbf{ASR} & \textbf{TTS} \\ 
		 & \textbf{WER(\%)} & \textbf{L2-norm$^{2}$} \\ \hline
		\multicolumn{3}{|c|}{\textbf{Baseline: ASR \& TTS}}\\ 
		\multicolumn{3}{|c|}{\textbf{(Supervised learning - Type 1)}}\\ 
		\hline
		$D^{P_{xyz}}_{xy}$2k\textsuperscript{*} & 81.31 & 0.874                               \\ \hline \hline
		\multicolumn{3}{|c|}{\textbf{Proposed: speech chain ASR$\rightarrow$TTS and TTS$\rightarrow$ASR}} \\
		\multicolumn{3}{|c|}{\textbf{(Semi-supervised learning - Type 2(a)\&2(c))}}\\ \hline
		+$D^{U_{xyz}}_{xy}$7k & 10.60 & 0.714                              \\ 
		\hline \hline
		\multicolumn{3}{|c|}{\textbf{Proposed: visual chain $\rightarrow$ speech chain}}\\
		\multicolumn{3}{|c|}{\textbf{Semi-supervised learning - Type 3(b)\&3(a))}}\\ \hline
		+$D^{U_{z}}$10k & 7.97 & 0.645                               \\ \hline \hline
		\multicolumn{3}{|c|}{\textbf{Topline: ASR \& TTS separately}}\\
		\multicolumn{3}{|c|}{\textbf{(Supervised learning - Full Data)}}\\
		\hline
		$D^{P_{xyz}}_{xy}$29k & 2.37 & 0.398                              \\ \hline 
	\end{tabular}
	
\end{table}
\vspace{-0.4cm}

\begin{table}[h]
	\vspace{-0.5cm}
	\centering
	\caption{IC and IR performance using Multimodal Chain\label{tab:icir}}
	\begin{tabular}{|l|c|c|c|}
		\hline
		\multicolumn{1}{|c|}{\multirow{2}{*}{\textbf{Data}}} & \textbf{IC} & \multicolumn{2}{|c|}{\textbf{IR}} \\
		 & \textbf{BLEU1} & \multicolumn{1}{|c|}{\textbf{R@10$\uparrow$}} & \textbf{med r$\downarrow$} \\ \hline
		\multicolumn{4}{|c|}{\textbf{Baseline: IC \& IR}}\\
		\multicolumn{4}{|c|}{\textbf{(Supervised learning - Type 1)}}\\
		\hline
		$D^{P_{xyz}}_{yz}$2k\textsuperscript{*} & 33.91 & 26.88 & 34                              \\ \hline \hline
		\multicolumn{4}{|c|}{\textbf{Proposed: visual chain IC$\rightarrow$IR and IR$\rightarrow$IC}} \\
		\multicolumn{4}{|c|}{\textbf{(Semi-supervised learning - Type 2(b)\&2(c))}}\\ \hline
		+$D^{U_{xyz}}_{yz}$7k & 42.11 & 28.14 & 31                            \\
		\hline \hline
		\multicolumn{4}{|c|}{\textbf{Proposed: speech chain $\rightarrow$ visual chain}}\\
		\multicolumn{4}{|c|}{\textbf{Semi-supervised learning - Type 3(c)\&3(a))}}\\ \hline
		+$D^{U_{x}}$10k & 43.08 & 28.44 & 30                            \\ \hline \hline
		\multicolumn{4}{|c|}{\textbf{Topline: IC \& IR separately}}\\
		\multicolumn{4}{|c|}{\textbf{(Supervised learning - Full data)}}\\
		\hline
		$D^{P_{xyz}}_{yz}$29k & 66.27 & 62.42 & 5                            \\ \hline
	\end{tabular}
\end{table}

\vspace{-0.5cm}

\subsection{Only a Small Amount of Paired Data Exists -\\ Utilizing Multimodal Chain}
As explained in Sec. \ref{sec:mmchain}, in this section we demonstrate how our proposed multimodal chain mechanism deal with the data composition written in Table \ref{tab:data-composition} where there are only 2k speech-text-image paired data, 7k speech-text-image unpaired data and 10k single modality data.

Note that in this research, the idea is not to show that the multimodal chain can outperform a baseline that was only trained with a small dataset. Instead, we want to identify how much we can improve performance when only unpaired data are available or even how much more improvement is possible when the required data are no longer available. We are interested to see how much we can improve the ASR performance when no more speech and text are available to train the model.

Table \ref{tab:asrtts} shows the ASR and TTS results from the scenarios in Section \ref{sec:mmchain}. First, we trained them on 2k paired data $D^{P_{xyz}}_{xy}$ as shown in the first block using the supervised training method. This initial model is then used in the next step as the speech chain component. We continued the training into the speech chain using $D^{U_{xyz}}_{xy}$ 7k data and achieved 10.60\% WER and 0.714 L2-norm$^{2}$. Finally, using the IC model that was trained semi-supervisedly through Type 2(a)\&2(c), we decoded the image-only $D^{U_{z}}$ dataset which enables it to be used in speech chain. By this way, we achieved about 2.6\% WER improvement over the original speech chain \cite{tjandra_schain1} that was only trained using the speech and text datasets. This result proved that the cross-modal data augmentation from the image modality into this speech chain is correlated positively with model quality. Our proposed strategies makes improvement of ASR and TTS possible, even without any speech or text data, with the help of a visual chain.

Table \ref{tab:icir} shows the IC and IR results from similar scenarios with the improvement from the speech chain. First, we did training using paired 2k data and achieved the baseline score shown in the first block. Next, we semi-supervisedly trained the IC and IR models in the visual chain mechanism, and produced over 8.2 BLEU1 improvement, 1.26 recall at 10 (R@10) improvement and 3 point improvement for the median r metrics. Finally, in the third block we show that the visual chain can also be improved using speech data, by the help of speech chain. There was about 1 point improvement in terms of BLEU for IC (high is good) and median r for IR (low is good). This result also implies that using our proposed learning strategy, the IC and IR model can be improved even without image and text datasets available. Therefore, we showed that it also works not only from image-to-speech modality, but also reversely. 

\vspace{-0.4cm}
\section{Related Works}
\vspace{-0.3cm}
Human communication is multisensory and involves several communication channels, including auditory and visual channels. Such multiple signals in different form are processed based on their characteristics, but later can be used for mutual augmentation. Moreover, the sensory inputs from several modalities share complementary behavior to ensure a robust perception of the overall information.

Over the past decades, several studies have integrated audio and visual cues to improve speech recognition performance. Within recent deep learning frameworks, Petridis et al. \cite{petridis_av} proposed one of the first end-to-end audiovisual speech recognition schemes. Another approach is the ``Watch, Listen, Attend, and Spell (WLAS)" framework \cite{chung2017lip}, which is an extension of the LAS framework \cite{chan2016listen} for speech recognition tasks that utilize a dual attention mechanism that can operate in three ways: over visual input only, audio input only, or both. Afouras et al. \cite{afouras2018deep} also proposed a deep audio-visual speech recognition to recognize phrases and sentences spoken by a talking face with or without audio.

Therefore, although the idea of incorporating visual information for automatic speech recognition (ASR) is basically not new, most approaches are usually done by simply concatenating the information that is inefficient to effectively fuse information from various modalities. Furthermore, these methods require that all of the information from these modalities be available altogether, which is often difficult. In contrast, humans process different modalities by different organs (i.e., ears for listening, mouths for speaking, eyes for seeing, etc.), which enables independent processing of such information while simultaneously opening room for augmenting each other.

Previously, a machine speech chain, based on sequence-to-sequence deep learning, was proposed to mimic speech perception and production behavior. It separately processed listening and speaking by ASR and a text-to-speech synthesis (TTS), but also enabled semi-supervised learning to teach each other when they received unpaired data. Unfortunately, our study is limited to speech and textual modalities. In fact, natural communication is actually multimodal because it involves both auditory and visual sensory systems

In this research, we take a further step to construct a multimodal chain and design a closely knit chain architecture that combines ASR, TTS, image captioning, and image production (retrieval or generation) models into a single framework. The framework allows each component to be trained without a large number of parallel multimodal data. Our experimental results show that further training of an ASR without available speech and text data remains possible by cross-modal data augmentation through our proposed multimodal chain, which improves ASR performance.

\vspace{-0.5cm}
\section{Conclusion}
\vspace{-0.3cm}
We described a novel approach for cross-modal data augmentation that upgrades a speech chain into a multimodal chain. We proposed a visual chain by jointly training IC and IR models in a loop connection that can learn semi-supervisedly over an unpaired image-text dataset. Then we improved the speech chain using an image-only dataset, bridged by our visual chain, and vice-versa. Therefore, we conclude that it is still possible to improve ASR, even without speech and text data available, with our proposed multimodal chain. We showed that each model in both chains can assist each other given an incomplete dataset by leveraging the data augmentation among modalities. In the future, we will jointly train both the speech and visual chain so that both can also be updated together. Furthermore, following the previous speech chain \cite{tjandra_schain2} that can synthesize multi-speaker speech, we will investigate our multimodal chain on natural multispeaker speech dataset as well.

\vspace{-0.6cm}
\section{Acknowledgements}
\vspace{-0.2cm}
Part of this work is supported by JSPS KAKENHI Grant Numbers JP17H06101 and JP17K00237 as well as NII CRIS Contract Research 2019 and Google AI Focused Research Awards Program.

\bibliographystyle{IEEEtran}
\bibliography{mthesis}

\end{document}